\documentclass[10pt,twocolumn,letterpaper]{article}

\usepackage[pagenumbers]{wacv} % To force page numbers, e.g. for an arXiv version

\usepackage{graphicx}
\usepackage{amsmath}
\usepackage{amssymb}
\usepackage{booktabs}

\usepackage[pagebackref,breaklinks,colorlinks]{hyperref}

\usepackage[capitalize]{cleveref}
\crefname{section}{Sec.}{Secs.}
\Crefname{section}{Section}{Sections}
\Crefname{table}{Table}{Tables}
\crefname{table}{Tab.}{Tabs.}

\def\wacvPaperID{2394} % *** Enter the WACV Paper ID here
\def\confName{WACV}
\def\confYear{2025}

\begin{document}

%%%%%%%%% TITLE - PLEASE UPDATE
\title{Rethinking Self-Supervised Learning Within the Framework of Partial Information Decomposition}

\author{Salman Mohamadi\\
West Virginia University\\
Morgantown, WV, USA\\
{\tt\small sm0224@mix.wvu.edu}
% For a paper whose authors are all at the same institution,
% omit the following lines up until the closing ``}''.
% Additional authors and addresses can be added with ``\and'',
% just like the second author.
% To save space, use either the email address or home page, not both
\and
Gianfranco Doretto\\
West Virginia University\\
Morgantown, WV, USA\\
{\tt\small gianfranco.doretto@mail.wvu.edu}
\and
Donald A. Adjeroh\\
West Virginia University\\
Morgantown, WV, USA\\
{\tt\small donald.adjeroh@mail.wvu.edu}
}
\maketitle

%%%%%%%%% ABSTRACT
\begin{abstract}
   Self Supervised learning (SSL) has demonstrated its effectiveness in feature learning from unlabeled data. Regarding this success, there have been some arguments on the role that mutual information plays within the SSL framework. Some works argued for increasing mutual information between representation of augmented views. Others suggest decreasing mutual information between them, while increasing task-relevant information. We ponder upon this debate and propose to revisit the core idea of SSL within the framework of partial information decomposition (PID). Thus, with SSL under PID we propose to replace traditional mutual information with the more general concept of joint mutual information to resolve the argument. Our investigation on instantiation of SSL within the PID framework leads to upgrading the existing pipelines by considering the components of the PID in the SSL models for improved representation learning. Accordingly we propose a general pipeline that can be applied to improve existing baselines.
Our pipeline focuses on extracting the unique information component under the PID to build upon lower level supervision for generic feature learning and on developing  higher-level supervisory signals for task-related feature learning. In essence, this could be interpreted as a joint utilization of local and global clustering. Experiments on four baselines and four datasets show the effectiveness and generality of our approach in improving existing SSL frameworks.
\end{abstract}

%%%%%%%%% BODY TEXT
\section{Introduction}
\label{sec:intro}

SSL is among the most efficient learning principles along with deep active learning \cite{mohamadi2022deep} and semi-supervised learning,that exploit the power of unlabeled data towards pre-training architectures for real world downstream tasks. The idea is to train a model to learn how to solve a pretext/proxy task while being supervised via a signal from unlabeled data, guided by a loss function corresponding to the given SSL framework. One interpretation of SSL \cite{chen2021exploring,li2020prototypical,bai2022directional} considers instance-wise SSL frameworks as performing K-means clustering on augmented views from a sample, i.e., assigning same centroid to the  views of the same sample. There have been rich explorations on two main components of SSL frameworks, i.e.,  developing better and more oriented pretext tasks as well as developing more effective loss functions. This resulted in the emergence of different types of baselines, which include contrastive \cite{chen2020simple}, non-contrastive \cite{grill2020bootstrap,chen2021exploring}, clustering-based \cite{caron2018deep,caron2020unsupervised}, hard/soft whitening (redundancy reduction) \cite{ermolov2021whitening,zbontar2021barlow},  etc.
Even though the performance of SSL frameworks are very promising, in justification of its theoretical formulations, there have  been arguments on how mutual information contributes to  SSL performance.
The conventional idea used to  be that a given framework needs to \textit{maximize} the mutual information between \textbf{representations} of augmented views from the same sample, upper bounded by the mutual information of the views themselves \cite{sordoni2021decomposed, cover2006thomas}. 

However, other related work \cite{rainforth2018tighter,tschannen2019mutual} cast doubt on the intuition that more mutual information means better representation. Moreover, authors  in \cite{tian2020makes}  resort to Info-Min principle to come up with the idea that augmentation should be adjusted to generate views with \textit{less} mutual information between them without reducing task-associated information. Another recent work \cite{sordoni2021decomposed} proposed to perform conditional estimation of mutual information relying on importance of mutual information between views.
Mutual information utilized in most prior approaches are  between \textbf{two variables (representations of two views)}. 
We propose to reformulate this as a three-variable information system, involving representation of two augmented views and one target representation. This leads us to redirect our attention from mutual information between representation of two augmented views, to the joint mutual information between the respective representation of the two views and the target representation that we 
wish to learn. This will resolve the arguments by providing a more general framework to study the SSL from the lens of information theory.

To this end, we propose to rethink SSL under the theory of the partial information decomposition (PID) for modeling the information system. This new perspective defines a more general measure, joint mutual information, and enables decomposing the joint mutual information into three components, namely,  unique, redundant, and synergistic components as defined by \cite{williams2010nonnegative}. Rethinking SSL from this PID viewpoint provides an opportunity for a significant improvement on expanding the supervisory signal for improved feature learning, regardless of the basic paradigm. In fact, pretext task and loss function as two building blocks of SSL have been well explored. However, the study of the role of supervisory signal in self-supervision has remained under-explored. Among redundant, synergistic, and unique information components of PID, only recently have  \cite{zbontar2021barlow,ermolov2021whitening} explored the redundancy component, while the other two components (synergistic and unique information components), are yet to be well-explored. Build on PID allows to resolve the argument on the role of mutual information, by providing a more general framework based on joint mutual information. We explore all three components of PID within SSL and develop a framework based on unique component of information.
Revisiting the intuition behind SSL from the PID perspective 
leads us to upgrade the existing SSL pipelines to accommodate all the three information components from PID, 
resulting in learning of more improved representations.  
% Augmentation is important as it quantitatively defines the interaction between source variables (two views), upgrading the existing SSL pipelines is important as it allows us to play with three other components (redundancy, synergy, and unique components) rather than just mutual information. In this work, we choose the second strategy, and build on this to develop a general SSL pipeline.

SSL is implemented mostly by contrasting the representation of two or more augmented views of a sample either with each other (positive pairs) or with those of other samples (negative pairs). From a reductionist point of view, as the augmentation process is cascaded with the SSL model, one can see supervision as an implicit binary supervisory signal inherent in the process of contrasting representations. In other words, considering a sample data, augmented views of this sample are implicitly labeled as though coming from the same sample (label 1), while augmented views of other samples are labeled differently (label 0). Now rethinking this pseudo supervision and comparing it with the original supervision in supervised learning, we argue that the fundamental difference is in the level of supervision. Here we only have a binary label every time we feed a sample to the model (say label 1 for all positive views, and label 0 for all negative views), whereas in supervised learning we enjoy more general and higher level labels such as class label in supervised classification setting. 
Viewing the SSL baselines from this perspective, we see that
these baselines somehow implicitly assign binary labels to augmented views and \textbf{do not go beyond sample level supervision} as they only consider the association between views in the level of sample (as opposed to class). Essentially, one important issue that recently gained attention is that this type of supervision is often implemented via rigorous invariance enforcement to data augmentation which turned out to be seriously harmful to downstream task \cite{huang2022learning}. \textit{Therefore, in supervised learning the supervision is a label for each sample, and as we have a finite set of labels, then samples share the supervision in a higher level (i.e., samples with the same class label), \textbf{whereas} in SSL frameworks we normally have assignment of implicit binary labels to the views of samples where even views from different samples \textbf{but the same class} get different labels.} We will show how PID allows for extending the supervision in SSL frameworks.  
Briefly, our key contributions in this work are as follows:
\begin{itemize}
    \item Rethinking SSL within the PID framework, to resolve the arguments on theoretical justifications regarding whether one should increase or decrease mutual information between representation of positive views.
    \item Designing a general pipeline based on the PID components to improve the existing    
    SSL models, which leads to expanding the self-supervision to progressive self-supervision.
    \item Experimental results on three benchmark datasets and detailed comparison with four widely used baselines to assess the effectiveness of the proposed pipeline. Results show the improvement offered by our framework.
\end{itemize}

\section{Related Literature and Preliminaries}
\label{lit} 
\textbf{{SSL}}:
SSL frameworks evolved mainly along with two components, pretext task and loss function \cite{mohamadi2023fussl}. While there was a need for task-specific networks in earlier SSL frameworks, introducing standard data augmentation in \cite{chen2020simple} eliminated the need for task specific network design. 
Loss functions also diversified depending on the intuition behind specific learning representations \cite{mohamadi2024active}. Due to lack of space, we elaborate on more details in supplementary materials.

\textbf{PID, and Joint Mutual Information vs Mutual Information}: 
The PID is a theoretical framework to model an information system with at least three variables (two source variables and one target variable) which allows decomposition of the joint mutual information regarding the target variable within the system into non-overlapping components \cite{williams2010nonnegative}. In fact, PID allows for generalization of conventional mutual information (between two variables) to joint mutual information between more than two variables \cite{mohamadi2023more}. The basic premise is that the source variables carry information about the target variable, and unlike challenging difficulties in decomposing the joint mutual information, the ideal was to perform a decomposition with non-negative information components. Finally, an earlier work in  \cite{williams2010nonnegative} proposed a method for non-negative decomposition of joint mutual information for information systems with finite variables. The proposed technique defines three components namely unique, redundant, and synergistic information components that are non-negative and could be separately quantified. Non-negative decomposition of the joint mutual information along with separate quantification of its components are among the advantages of PID \cite{timme2014synergy}.

Let's consider the simplest case, a system of three variables, two source variables $S_1$ and $S_2$ forming joint mutual information $I(S_1,S_2; T)$ with respect to a target variable $T$. The interaction between the information of the variables gives 
%birth 
rise to joint mutual information between them. However, a longstanding problem across multiple problem domains, is how to perform a decomposition allowing to pinpoint the contribution offered by each source variable and combination of sources about the information of the target variable. As formulated in \cite{williams2010nonnegative}, here the joint mutual information between sources and target, is made of three elements, unique, redundant, and synergistic information components.  Within the interaction of the variables, the part of information only provided by each source variable separately is called unique information, whereas redundant and synergistic information are defined around the both source variables. Redundant information which is also known as common mutual information is defines as the minimum information provided by each source, whereas synergistic information is the information provided only by a combination of two source variables about the target variable, which is not accessible using one source variable alone \cite{gutknecht2021bits}.
\begin{equation}
    \label{eq.PID}
    \scriptsize
    \begin{split}
         I(S_1,S_2:T)=\text{R}(T; S_1, S_2) + 
         \text{Sy}(T; S_1, S_2)+ \text{U}(T; S_1) + \text{U}(T; S_2)
    \end{split}
\end{equation}

where $R$, $Sy$, $U$ are redundant, synergistic and unique components of information respectively.
We believe that this information system modeling is much richer than the common mutual information and we adopted it as we find it better fit to model SSL. In this regard, considering one of the most general setting of SSL frameworks, we have two randomly augmented views generated for a sample, and a model to contrast their encoded representation toward learning the target representation, distribution of the samples. Here is exactly where we wish to argue that the modeling of interaction of augmented views could open three parallel windows (unique, redundant, and synergistic elements) to improve the performance, if done withing the PID framework. The represenation of two augmented views are considered as our source variables, while the original data distribution is the target variable $T$. 
Now regarding the arguments on whether increase or decrease mutual information, the conventional interpretation is that SSL approaches maximize the mutual information between deep representation $f(.)$ of two augmented views, $x_t$ and $x_{t'}$ for any given original sample $x$, which submit to the inequality $I(f(x_t);f(x_{t'}))\leq I(x_t;x_{t'})$ \cite{sordoni2021decomposed, cover2006thomas}. This is along with the idea that we want the learned representation to be maximally informative of data sample distribution and minimally representing the augmentation/noise effect. \textbf{However, from the perspective of PID, the question no longer is whether to increase or decrease mutual information, rather, we propose to upgrade a given SSL framework with respect to PID in order to reduce the redundancy, and increase the synergy as well as unique information}. More foundation on PID is available in Supplementary. 

\section{Progressive supervision via PID}
%using  unique information component}

Revisiting the general setting of SSL from the perspective of PID, replace the question of "whether we need to increase or decrease mutual information" to how implement SSL systems efficiently with regard to the three PID components of information toward unlocking the full potential of unlabeled data. Hence, in order to learn the target representation, one can play with three components, synergy, redundancy and unique information. It is very important to note that among these three information components,  quantification of two of them, synergy and redundancy, requires two representations (two source variables), while quantification of the unique information needs only one representation. Accordingly, the existing approaches incorporate only two elements, synergy and redundancy, that require two source representation to estimate the target representation. In other words, the reason is because these frameworks contrasts at least two positive views, hence they tend to learn the target representation from two source representations. Reducing the redundant component of information is recently studied in \cite{zbontar2021barlow,ermolov2021whitening}. We provide an 
%present a thorough 
investigation of synergy and redundancy within recent SSL frameworks in supplementary materials. 

%However, 
The main focus of this work is on exploiting the unique component of information as defined under PID, which is missing in current frameworks. This can involve modifying the contrastive learning objective to encourage the extraction of features that capture unique information about the individual views. In this regard, the goal is to extend current SSL pipeline to utilize the unique information in addition to two other components. We hypothesize that another level of self-supervision is needed in the process. i.e., one supervisory signal can not properly pinpoint all three components. Hence we motivate our pipeline as shown in Fig. \ref{Fig1} by emphasizing on progressive supervision as it is build upon basic sample level supervision of a given SSL framework toward progressing the supervision to higher level supervision. In fact, current self-supervision in SSL only provides associations between views of the same sample, while we seek a self-supervision that can associate views of different samples (but same cluster). 
Hence, given a SSL framework to be upgraded, here the core idea is to initially train the given model, and then leverage the learned features to label the original samples by clustering them using 
$k$-means$++$ algorithm on the output of the initialized model. Then the training enters another phase in which we aim at extracting unique information from each augmented view representing the unique information of PID using pseudo-labels. In this phase, we have an updated loss function designed to achieve two goals simultaneously: 
\begin{itemize}
    \item \textbf{Low Level Self-Supervision:} Enforce the invariance to the representation of augmented views via an SSL loss function, which enables the model to learn the features via a supervisory signal guiding at the level of samples. It can be interpreted as a local clustering.
    \item \textbf{ High Level Self-Supervision:} Enforcing the invariance to representation of views from samples with the same pseudo-label, i.e., coming from the same cluster/class. This is where the higher level supervision (class/cluster supervision) comes into play. It can be interpreted as a global clustering.
\end{itemize}
Note that this phase is performed in an iterative manner, meaning that after a certain number of epochs, the labeling will be renewed using $k$-means$++$ clustering.  This is where \textbf{progressive supervision} is instantiated, i.e., as training progresses, the model is optimized with both types of aforementioned supervisory signals, while the second one (pseudo-labels) gets updated progressively. Taken all together, in the upgraded pipeline, we have all three information components of PID, including once missing unique information. This is primarily due to the interplay between mentioned two types of supervision. 
In the following subsections, first we formalize SSL under PID, and then elaborate on each phase of training, their type of supervisory signal, and the loss function.

% \textbf{Supervision in SSL:} We start our contemplation by contrasting the type of supervision in existing SSL baselines with the original supervision in supervised learning. As discussed before, almost all of SSL baselines regardless of their underlying principles, exploit the supervision from sample itself, as the augmentation process is pinned to the rest of the framework, views of a sample are tagged as siblings as they are mostly simultaneously being processed. Even though the literature suggests that this supervision allows for learning general features, we argue that this is fundamentally different than supervised learning. In fact the type of supervision in supervised learning tasks such as classification is higher level as all samples from the same class are assigned the same label whereas in SSL the model only needs to distinguish the views from sample at hand from negative views. In classification task, the training process is guided by labels so that the model extract the information from each sample individually. In this sense, the availability of labels allows the learning process to target the unique information in each sample, needless of contrasting views or samples. 
\subsection{SSL under PID}
Here we depict the unique aspects of SSL based on the PID framework. To instantiate three components of PID within the SSL, for the sake of simplicity, we consider a three variable PID system, i.e., a SSL framework that generates two views for each sample. Suppose a SSL framework generates two views ($x_1$ and $x_2$) for any given sample $x$ from a sample set, to estimate the representation of samples, $T$. The SSL framework generates representation of each view, where the corresponding representations for a sample is a random vector (or tensor) $V= \{V_1, V_2\}$ carrying information about the target representation $T$ as a random variable. The goal is to decompose the information provided by views' representation vector $V$ about target representation $T$, to quantify the partial information offered by subsets of $V$ ($\{V_1\}$ and $\{ V_2\}$) individually or jointly in terms of unique, redundant, and synergistic information. Roughly speaking, the SSL framework is designed in such way that contrasts the representations of views and draw the representation of views from the same sample together. The information interaction between representations of views within the SSL framework is resulted from a direct or indirect contrast of representations. Accordingly the information interaction enables the existing SSL framework to learn the target representation using only synergistic and redundant information, which are information components coming from the two representation $V_1$ and $V_2$ and not just one of them. The unique information from each representation however, is missing within a typical SSL framework as this representation learning only learns the target representation by contrasting two other representations, not one. In fact synergistic and redundant information are components that in order for model to capture them, it \underline{requires two representations and not one}, whereas unique information is to be captured separately from each view, such as the case with supervised learning where the model captures the information of each sample solely based on its label (as opposed to based on contrasting it with other samples). Recently, some work on SSL based on hard and soft whitening \cite{zbontar2021barlow,ermolov2021whitening} have developed redundancy reduction-based approaches in representation learning. More detail on practical ways to capture redundant and synergistic information is available in supplementary.   
\begin{figure*}
%\label{Fig1}
  \centering
  \includegraphics[scale=.31]{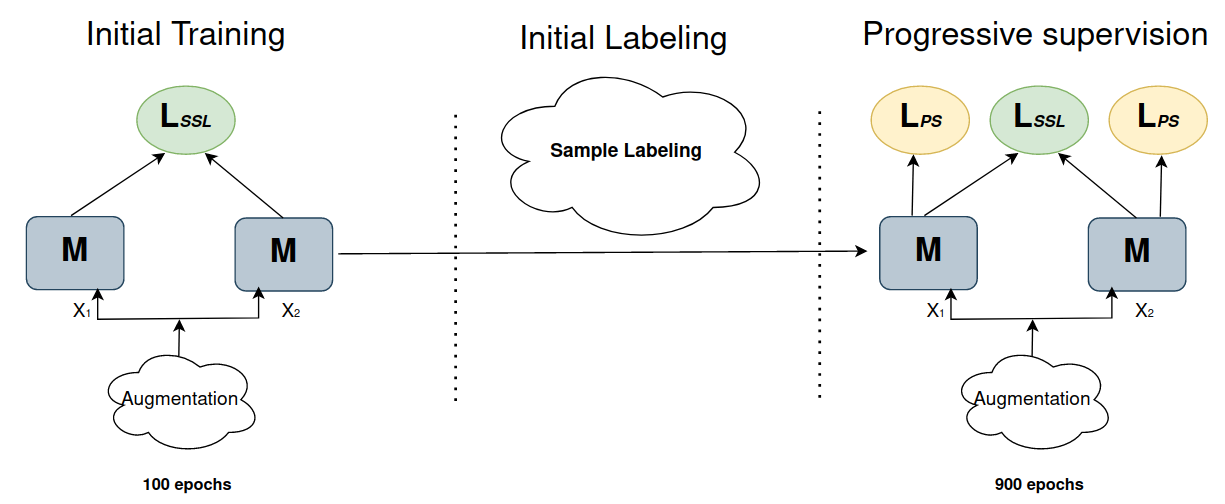}
		\vspace{-0.1in}
		\caption{ \scriptsize
		A given SSL framework undergoes an initial training, and clustering-based labeling, before the main phase of training, \textbf{iterative refinement}. The three-variable information system under PID framework is instantiated to have  all three components of PID, namely unique, redundant and synergistic components. The detailed derivation supporting this design is depicted in Supplementary. Our framework differentiates itself from "paradigms including clustering in SSL" by \textbf{joint invariance enforcement} to augmented representations, toward learning associations of views at both sample level and cluster level. The two components of PID, synergistic and redundant information are instantiated via iterative sample-level representation learning where the two views are \textbf{jointly} involved, whereas the third component of PID, unique  information, is instantiated via iterative supervised training using pseudo-label for each individual view.In essence, this framework allows for joint local and global clustering.}

\label{Fig1}

\end{figure*}

\subsection{Initial training}
The aforementioned unique component of the information defined by PID is missing in traditional SSL frameworks. To capture and utilize that for self-supervised representation learning, we need a framework that goes beyond contrasting two views, and improve the learned target representation using the representation of each view individually.
The proposed SSL framework starts the training process with initial training for some $N$ epochs in order to enable the framework to learn some features which are essential to initiate next phase of training. This generates the supervisory signal operating at sample level, enforcing invariance to the representation of augmented views of each sample. Hence, it  does not relates negative views (views from different samples) even if they come from the same class. From the perspective of three variable information system within PID, as the training is mainly performed on the contrasting views, we mentioned that the initial training only directly focuses on redundancy (such as \cite{ermolov2021whitening,zbontar2021barlow} and synergy, and not unique information offered by each view individually. 
During the completion of initial training the model learns some useful features which could be used to distinguish the samples through clustering. Later we will leverage this to enables clustering and labeling (pseudo-labeling) samples at the next step in a joint learning process. %Regarding 
For loss function, here we only have the SSL loss function, $\mathcal{L}_{ssl}$, associated with the given SSL framework that is meant to be upgraded.

\subsection{Pseudo-labeling}
Initial training of a given framework enables the model to learn and retain some general features as SSL frameworks aim at learning meaningful representation covariant with the augmentation techniques. These features are mostly expected to be low to medium level features which would be useful for clustering. We perform a clustering technique, $k$-means$++$ algorithm \cite{arthur2007v} on the output of the projector of the given framework, fed by the original samples to leverage the learned features to cluster them into a certain number of clusters depending on the prior knowledge of number of classes of the dataset. These labeled (hard pseudo-labeled) samples will be used for next phase of training.  

\subsection{Progressive self-supervision} 
The progressive self-supervised training follows the pre-training and pseudo-labeling, which fully instantiate the PID information components in SSL. Specifically, this phase of training integrates all three information components of PID (especially unique information) as it is equipped with double supervision. That is,  using two types of supervisory signal, namely, sample-level supervision and cluster-level supervision. 
%, where the 
The latter one is built upon the former phase of training, and is going to be progressively updated each time, after a certain number of epochs via $k$-means$++$ re-clustering. Note that the pseudo-labels get updated at each iteration. The core idea is to jointly learn an invariant representation for views from the same sample as well as views from the same cluster. 
As shown in Fig. \ref{Fig1}, the second phase of training utilizes the pseudo-labels via $\mathcal{L}_{PS}$, simultaneously along with sample supervision via $\mathcal{L}_{SSL}$. Specifically, in the second phase of training, for a given sample $x$, two augmented views $x_1$ and $x_2$ are generated. The loss function encourages two objectives, learning invariant representation for $x_1$ and $x_2$ via $\mathcal{L}_{SSL}$, as well as learning the right cluster/class via classification loss, $\mathcal{L}_{PS}$ supervised by a progressive supervisory signal. The total loss for a given SSL framework with $\mathcal{L}_{SSL}$ is denoted as follows:
\begin{equation}
    \label{eq-}
    \centering
    \scriptsize 
    \begin{split}
  &\mathcal{L}=\mathcal{L}_{SSL}(z_1,z_2)+\alpha\{\mathcal{L}_{PS}(z_1, \hat{y})+\mathcal{L}_{PS}(z_2, \hat{y})\}\\
  & \mathcal{L}_{PS}(z, \hat{y}) = - \sum_{c=1}^C \hat{y}_{c}\log (\text{softmax}(z_c))\\
    \end{split}
\end{equation}
where $\mathcal{L}_{PS}$ is classification loss, $z_1$, and $z_2$ are the respective outputs of projectors for two augmented views, $\hat{y}$ is the pseudo-label for both views of a sample, and $\mathcal{L}_{SSL}$ is the given SSL loss function.  Moreover, $C$ denotes the number of classes, while $\hat{y}_c$ is the ground truth probability of class $c$, and finally softmax($z_c$) is the predicted probability of class c after applying the softmax function to the output z. Specifically, $\mathcal{L}_{PS}$ enables the model \textbf{to learn representation via extracting the unique information of each view (unique component of PID)} as it learns from the classification of views individually and independently. $\alpha$ is a weighting factor that grows as the training progresses. As the initial iterations of labeling are less accurate, %very small amount of $\alpha$ at the 
we start with small values for $\alpha$ at the first few iterations of labeling, which allows \underline{retaining} the learned features from lower level supervision (invariance enforcement to the representation of positive views, $\mathbf{L}_{SSL}$). As the training advances, the growth of $\alpha$, gives more weight to the higher level supervision (enforced by classification loss $\mathcal{L}_{PS}$) which tends to get more accurate as the training process progresses. 
In fact, the pseudo-labels provided by clustering at each iteration enables treating the augmented views that come from different samples but belong to the same cluster the same way. {In the long run, this aims at creating a more compact collective representation of samples from the same cluster, by pulling representation of samples from the same cluster together while pushing samples from other clusters further apart in an iterative refinement procedure.}

% Mathematically, the optimal solution to capture unique information of each sample would be:

% \[
% z_{i,c} = \begin{cases} 
%   \text{Large positive value} & \text{if } c = \text{argmax}(\hat{y}) \\
%   \text{Small negative value} & \text{otherwise} 
% \end{cases}
% \]

% In this scenario, the softmax function applied to \(z_{i,c}\) would yield a probability close to 1 for the true cluster label and close to 0 for other labels, resulting in a minimal loss.

% This demonstrates that the pseudo-labeling loss encourages the model to learn representations that align with the unique information contained in each view, as the optimal solution involves assigning high probabilities to the true cluster label for each view. Therefore, our framework effectively captures the unique component of PID by encouraging the extraction of unique information from individual views through the pseudo-labeling loss while the SSL loss involves redundant and synergistic information.
% In the next section we show the alignment of the framework with Information Bottleneck (IB) principle

\subsection{Alignment with Information Bottleneck (IB) principle }
 Given the IB princinple, we want to show that our total loss function is aligned with the IB principle at two levels: "intra-alignment" and "inter-alignment", viz:
 
 % Let's define "intra-alignment" and "inter-alignment" in the context of loss function alignment with the IB Principle:

\textbf{Intra-alignment:}
Intra-alignment refers to the alignment between a single term in our loss (first term) and the IB Principle. Specifically, for a SSL loss to be considered intra-aligned with the IB Principle, it should encourage the model to learn a bottleneck representation that balances the trade-off between canceling the augmentation effect (mutual information between learned representation $T_\theta$ and augmentation $X$) and learning the target representation (mutual information between learned representation $T_\theta$ and target representation $Y$). Here,  $\theta$ denotes the weights of the model. Mathematically, this can be expressed as:
\begin{equation}
    \label{eq-IB}
    \centering
    \scriptsize
    \begin{split}
     % \min_{p(z)} {IB} \triangleq \min_{p(z)} \left( I(Z;Y)-\beta I(Z;X)\right)\\
      \min \mathcal{L}_{SSL} \approx \min {IB_\theta} \triangleq \min \left( I(X;T_\theta) - \beta \cdot I(T_\theta;Y)\right);
    \end{split}
\end{equation}
where \( \mathcal{L}_{SSL} \) is the SSL loss, \( I(X; T) \) represents the mutual information between input augmentation effect $X$ and learned representation \( T \), \( I(T; Y) \) represents the mutual information between representation \( T \) and data distribution or target representation \( Y \), and \( \beta \) controls trade-off between canceling the augmentation effect and learning the target representation. Our framework takes a valid SSL framework and updates it, according to PID information system. 
%, therefore, Because the
Thus, since the first term of our loss function, $\mathcal{L}_{SSL}$ is already taken from a valid SSL framework that follows the IB principle (e.g., Barlow-Twins, SimCLR, ...), the first term already satisfies the intra-alignment with IB principle.
%(for each framework, please refer to the original framework for mathematical proof).
Note that given that the pseudo-labels are accurate, the second term in the loss, need not proof of alignment with IB as it is in fact supervised learning. 

\textbf{Inter-alignment:}
Given that the type of supervision in supervised learning (using the ground truth labels) which is ideal supervision for learning the representation, the standard mechanism to substitute this supervision in SSL frameworks relies on a pseudo-supervision (supervisory signal) that generates sort of implicit sample-level labels (which associates views of the same sample, as opposed to ground truth labels that associates the samples globally). Accordingly, the learned representation via a SSL framework, should ideally allows for prediction of ground truth labels. However, given a typical SSL framework that operates on enforcing invariance to the representation of augmented (distorted) views of the same sample, one shortcoming is that this framework does not involves (or even might undermine) the associations beyond views of the same sample. In other words, since there is no access to the labels, it does not treat the samples with the same ground truth label as if they are from the same representation. Taken all these together, our framework, allows from involving such association using incorporation of unique component of information.
Inter-alignment refers to the alignment between different terms within our total loss function and the IB Principle. For the entire loss function to be inter-aligned with the IB principle, the combination of terms should collectively balance mutual information between sample-level association effect  $X'$ and learned representation $T'$, and the mutual information between $T'$ and global (class or cluster level) sample association $Y'$. %Mathematically, IB principle can be framed as:
Following the IB principle, this can be framed as: 
\begin{equation}
    \label{eq-IB}
    \centering
    \scriptsize
    \begin{split}
     % \min_{p(z)} {IB} \triangleq \min_{p(z)} \left( I(Z;Y)-\beta I(Z;X)\right)\\
      \min {IB_{\theta'}} \triangleq \min \left( I(X';T'_{\theta'}) - {\beta'} \cdot I(T'_{\theta'};Y')\right);
    \end{split}
\end{equation}
where  $\min I(X';T'_{\theta'})$ comes down to proper minimization of the SSL loss term, and $\min( - {\beta'} \cdot I(T'_{\theta'};Y'))$ comes down to proper minimization of the pseudo-labeling loss term, and \( {\beta'} \) is a weighting factor.

Inter-alignment ensures that the combination of terms in the loss function collectively encourages the model to find an optimal bottleneck representation.
Specifically, the total loss allows for learning a representation that is similar to that of supervised learning due to second term while the unwanted effect of invariance enforcement to augmentation through SSL loss is compensated. 

% Hence given
% \begin{equation}
%     \label{eq-IB}
%     \centering
%     \begin{split}
%      % \min_{p(z)} {IB} \triangleq \min_{p(z)} \left( I(Z;Y)-\beta I(Z;X)\right)\\
%        \min {IB} \triangleq \min \left( I(X;T) - \alpha \cdot I(T;Y)\right); \mathcal{L}_{SSL} = I(X;T); \mathcal{L}_{PS} = -I(T;Y)
%     \end{split}
% \end{equation}

% % \[ \mathcal{L}_{SSL} = I(X;T) \]
% % \[ \mathcal{L}_{PS} = -I(T;Y) \]

% By combining the two loss components with the weighting factor \( \alpha \), the loss function achieves a balance between first term (SSL loss: invariance enforcement to representation of views from the same sample) and second term (supervised learning classification loss). 

\section{Experiments and results }
\vspace{-0.12cm}
In this section, we evaluate the effectiveness of proposed approach and demonstrate its versatility by integrating it into multiple widely used SSL models. This includes a set of $K$-nn evaluation, supervised linear evaluation, as well as transfer learning on a number of datasets\footnote{We are thankful to a newly published SSL library, Solo-Learn \cite{da2022solo}, for providing the implementation of all SSL baselines.}. \\
\textbf{Baselines:} In order to assess the effectiveness of our pipeline, we incorporate most recent strong baselines into comparison, covering contrastive, non-contrastive, and whitening baselines, which include: SimCLR \cite{chen2020simple}, BYOL \cite{grill2020bootstrap}, Whitening-MSE ($d=2$) \cite{ermolov2021whitening} and B-Twins \cite{zbontar2021barlow}.\\
\textbf{Datasets:} Four datasets are used in this study, including ImageNet \cite{deng2009imagenet}, CIFAR10 \cite{krizhevsky2009learning},  CIFAR100 \cite{krizhevsky2009learning}, and Tiny ImageNet \cite{le2015tiny}.

% CIFAR10 and CIFAR100 are small size datasets consisting of 60k images of dimensions $32\times\ 32$ , with 50k training samples and 10k testing samples, in 10 and 100 classes respectively. 

% Tiny ImageNet \cite{le2015tiny} is a smaller version of ImageNet (ILSVRC2012) made of over 100k samples of dimension $64 \times 64$ in 200 classes.

\subsection{Experimental setting}

\textbf{Architecture:}
For CIFAR10 and CIFAR100 datasets, the encoder is built with ResNet18, whereas for ImageNet and Tiny ImageNet the encoder architecture adopts ResNet50. In both cases the classification layer of ResNet is replaced with a three-layer projector as described in \cite{zbontar2021barlow} when applicable in the assessed baseline. During the second phase of training, the output of the projector is sent to a linear classifier (softmax preceded with fully connected layer) for classification.

\noindent\textbf{View generation details:} Augmentation protocols can be categorized into standard and heavy augmentation operations following recent works \cite{grill2020bootstrap,chen2020simple}, with standard augmentation protocol more widely used \cite{bai2022directional}. %In this work, 
We adopt standard augmentation operations proposed in~\cite{chen2020simple}. For each sample image in the four datasets, augmented views are generated using a randomly selected augmentation function $\tau$. $\tau$ consists of a set of operations including random crop, color jittering, aspect ratio adjustment, Gaussian blurring, horizontal mirroring, and gray scaling. \\
% Random crop between 0.2-1 as well as 0.08-1 of the original image size is performed on samples of  CIFAR10/100 and Tiny ImageNet respectively. Aspect ratio adjustment ranging from 3/4 to 4/3 image aspect ratio is performed. Horizontal mirroring, gray scaling, and color jittering each with respective probabilities of 0.5, 0.1, 0.8  is used in view generation as specified in \cite{chen2020simple} in detail.\\
\textbf{Implementation:}
We use Adam optimizer \cite{kingma2014adam} for model optimization in all the experiments (for both supervised training and testing).
Following original implementation of SimCLR \cite{chen2020simple}, we set $\tau=0.1$.
$\lambda$ is set to $5\times 10^{-3}$ adopting the suggested setting in B-Twins. In case of W-MSE \cite{ermolov2021whitening}, we adopt the setting of W-MSE2, generating two views for each sample. As a standard practice \cite{ermolov2021whitening}, the latent space of the given baseline is norm-two normalized whenever applicable. We perform evaluation in two different settings, namely supervised linear and transfer learning. In case of supervised linear evaluation of CIFAR10/100, the pre-training and evaluation is performed using ResNet18 as the encoder. And for transfer learning, we adopt ResNet50 as encoder pre-trained on ImageNet. We follow the details of \cite{chen2020simple} in case of transfer learning with ResNet50 on CIFAR10/100 datasets. The weight decay is set to $10^{-6}$ for all experiments.

\noindent\textbf{Initial training}
We perform initial training for 100 epochs on all baselines (i.e., B-Twins, BYOL, etc), adopting the same setting and pipeline suggested by the original work.  It starts with a learning rate of 0.1 for 20 epochs and switches to a learning rate of $10^{-3}$ for the remaining epochs. Take B-Twins as an example, here the twin networks are trained as suggested in \cite{zbontar2021barlow} except for only 100 epochs. 

% \noindent\textbf{K-means$++$ clustering}
%As mentioned previously, initial training of a given approach enables the model to generate representative features as SSL frameworks aim to learn meaningful representation covariant with the augmentation techniques. After initial training finishes, the weights are fixed and the original samples are fed to the model. $K$-means$++$ clustering is performed on the model output (for instance, in case of B-Twins, the output of the projector of one of the twin networks) to cluster the samples into $N$ clusters, and hard pseudo-labels are assigned accordingly. The labels are utilized for progressive SSL with the generated views at the next phase of training (with augmented views). Note that $N$ is chosen based on the prior knowledge of the dataset, i.e., $N$ is the underlying number of classes for each dataset. We also use this technique to re-assign updated labels in each iteration (each iteration $=$ 100 epochs) of the second phase of training.  

\noindent\textbf{Progressive self-supervision}
The main phase of training starts after initial training and pseudo-labeling, under a different setting as shown in Fig. \ref{Fig1}. This phase of training proceeds for 900 epochs. We fix the weights and utilize $k$-means$++$ to generate new pseudo-labels for every 100 epochs, which will be used for the next iteration of 100 epochs. This progressively improves the quality of higher level supervisory signal. The learning rate is set to $10^{-3}$. Hyperparameter $\alpha$ increases along with training iterations. Specifically, $\alpha=$ $10^{-5}$, $10^{-4}$, $10^{-3}$, $10^{-2}$, and $10^{-1}$ corresponding to training epochs of 101-200, 201-400,401-600, 601-800, and 801-1000 respectively. 
\subsection{Evaluation}
We follow standard practice of evaluating SSL frameworks \cite{goyal2019scaling,zbontar2021barlow,ermolov2021whitening,chen2020simple} for image classification task. The standard evaluation consists of removing the projector head and placing a trainable linear classifier (single fully connected layer followed with softmax layer) on top of fixed encoder, to be trained and tested in supervised manner with labeled evaluation data. We also perform newly emerged evaluation practice introduced in \cite{ermolov2021whitening}, in which deterministic classifier ($K$-NN with $k=5$) is used without further training for downstream tasks. In fact, as a useful practice along with standard evaluation, is $k$-NN evaluation that is also performed by some work such as \cite{ermolov2021whitening} which implies that pre-trained SSL models would enables classification of samples without further supervised training on the labeled data. \cite{ermolov2021whitening} directly tests SSL pre-trained models after self-training completion (such as 1000 epochs) with no further supervised training on the evaluation data.
\begin{table*}
\vspace{-1.5em}
  \centering
  \scriptsize
  % \footnotesize
  \begin{tabular}{p{1.1cm} p{01.cm} p{01.6cm}| p{01.cm} p{01.6cm}||p{01.cm} p{01.6cm}| p{0.9cm} p{01.5cm}} 
    \toprule
    % \multicolumn{2}{c}{Part}                   \\
    % \cmidrule(r){1-2}
    Framework   & \multicolumn{4}{c}{CIFAR10} & \multicolumn{4}{c}{CIFAR100}   \\
    \cmidrule(r){2-9}
    % \midrule
        & Original  &  Ours &  Original (Knn) & 
 Ours (Knn) & Original  &  Ours &  Original (Knn) &  Ours (Knn) \\
    \cmidrule(r){2-9}
      BYOL  &   91.81  & 93.12  & 89.39  &  91.77 & 70.47   & \textbf{72.40} {\tiny \textcolor{blue}{(+1.93)}} & 57.31 &   59.54   \\
       SimCLR  &  91.93 & 93.27  & 88.63  & 91.03 {\tiny \textcolor{blue}{(+2.4)}} &  {66.21}  & 67.91  & 56.55  & 57.91   \\
   {WMSE2}  & {90.16} &  {92.03}  {\tiny \textcolor{blue}{(+1.87)}}  & {88.93}    & 90.18 & {65.49} &  66.51 & {56.90}    & 58.56 \\
   {BTwins}  &    \textbf{92.55} & \textbf{93.97}   & \textbf{90.44} &  \textbf{92.59}  &  \textbf{70.79}  & 72.52   & \textbf{59.11} &    \textbf{61.57} {\tiny \textcolor{blue}{(+2.46)}}  \\
    \bottomrule
  \end{tabular}
  \vspace{-0.3cm}
  \caption{
  \scriptsize
  Top-1 classification accuracy under linear evaluation and $K$-nn ($K$=5) evaluation for CIFAR10 and CIFAR100. 
  %; as presented our framework upgrades 
  Our framework improves the results of all for baselines. Results under the $K$-nn classifier (without  supervised training) show the maximum improvement of the work, implying that our framework allows for learning of higher level (task-related) features, thanks to the progressive supervision. 
  %Note that for comparison purposes, 
  "Original" and "Original ($Knn$)" denotes the linear and $K$-nn evaluation results of the corresponding baseline with same number of epochs (1000 epochs) as its regular training.  
  %Top-1 classification accuracy under linear evaluation and $K$-nn ($K$=5) evaluation for CIFAR100, compared to CIFAR10, the results show a bit less improvement overall.
}
  \label{table1}
\end{table*}
% \vspace{-3em}

% \begin{table}
%   \label{table2}
%   \centering
%   \scriptsize
% %   \footnotesize
%   \begin{tabular}{p{1.cm} p{01.cm} p{01.5cm} | p{01.5cm} p{01.2cm}} 
%     \toprule
%     % \multicolumn{2}{c}{Part}                   \\
%     % \cmidrule(r){1-2}
%     Framework   & \multicolumn{4}{c}{CIFAR100}    \\
%     \cmidrule(r){2-5}
%     % \midrule
%         & Original  &  Ours &  Original (Knn) & Ours (Knn) \\
%     \cmidrule(r){2-5}
%       BYOL  & 66.60   & \textbf{68.83} \textcolor{blue}{(+1.23)} & 56.91 &   58.81 \textcolor{blue}{(+1.9)}   \\
%        SimCLR  &  \textbf{66.86}  & 67.79  & 56.14  & 57.55 \\
%     {W-MSE2}  & {66.15} &  67.26 & {56.63}    & 58.14 \\
%    {B-Twins}  &  66.79  & 67.81   & \textbf{57.34} &    \textbf{59.01}    \\
%     \bottomrule
%   \end{tabular}
%   \caption{
%   % \footnotesize
%  Top-1 classification accuracy under linear evaluation and $K$-nn ($K$=5) evaluation for CIFAR100, compared to CIFAR10, the results show a bit less improvement overall.
% }
% \vspace{-0.3cm}
% \end{table}
\begin{table*}
  \centering
\vspace{-0.3cm}
  \scriptsize
  % \footnotesize
  \begin{tabular}{p{1.cm} p{0.9cm} p{01.37cm} | p{0.9cm} p{01.4cm}|| p{0.9cm} p{01.4cm} | p{0.9cm} p{01.4cm}} 
    \toprule
    % \multicolumn{2}{c}{Part}                   \\
    % \cmidrule(r){1-2}
    Framework   & \multicolumn{4}{c}{ImageNet} & \multicolumn{4}{c}{Tiny ImageNet}  \\
    \cmidrule(r){2-9}
    % \midrule
       & Original  &  Ours &  Original (Knn) & Ours (Knn)  & Original  &  Ours &  Original (Knn) & Ours (Knn) \\
    \cmidrule(r){2-9}
      BYOL  & 74.2   & \textbf{75.7 } & 55.3   &  58.2  & \textbf{51.16} & 52.25 & \textbf{36.39} &  \textbf{39.08} {\tiny \textcolor{blue}{(+2.69)} } \\
       SimCLR  & 69.5  &  70.9 &  54.1  & 57   & 48.91  & 50.36 & 33.11 & 35.29  \\
    {WMSE2}  & 73.3 &  74.9 &  54.9  &  57.7  &  48.51 & 49.93 & 34.24 &  36.91  \\
   {BTwins} &  73.4 & 75.1 {\tiny \textcolor{blue}{(+1.7)}} &  55.1  & \textbf{58.4} {\tiny \textcolor{blue}{(+3.3)}}   & 50.87 & \textbf{52.44} {\tiny \textcolor{blue}{(+1.57)}} & 36.11 & 38.64    \\
    \bottomrule
  \end{tabular}
  \vspace{-0.3cm}
  \caption{
  \scriptsize
 Top-1 classification accuracy under linear evaluation and $K$-nn ($K$=5) evaluation for ImageNet and Tiny ImageNet.
}
  \label{table3}
\end{table*}

\begin{table}
  \centering
  \scriptsize
%   \footnotesize
  \begin{tabular}{p{1.cm} p{01.cm} p{01.37cm}| p{01.cm} p{01.37cm}} 
    \toprule
    % \multicolumn{2}{c}{Part}                   \\
    % \cmidrule(r){1-2}
    Framework   & \multicolumn{2}{c}{CIFAR10}   & \multicolumn{2}{c}{CIFAR100}  \\
    \cmidrule(r){2-5}
    % \midrule
        & Original  &  Ours & Original & Ours \\
    \cmidrule(r){2-5}
      BYOL  & 93.14 & 94.19  & 78.20  &  79.45    \\
       SimCLR  & 91.59  & 92.79  & 76.66  &   77.81   \\
    {WMSE2}   & 92.55 & 93.67   &  78.21 &  79.65    \\
   {BTwins}  & 94.47 & \textbf{95.82} {\tiny  \textcolor{blue}{(+1.35)}}  & 79.91 &   81.43 {\tiny \textcolor{blue}{(+1.52)}}   \\
    \bottomrule
  \end{tabular}
  \vspace{-0.3cm}
  \caption{
  \scriptsize
 Top-1 classification accuracy under transfer learning evaluation for CIFAR10/100, using ResNet50 encoder pre-trained on ImageNet, the results for W-MSE is using two views.
}
  \label{table4}
\end{table}

%\vspace{-0.5em}
\subsection{Results}
\textbf{Linear evaluation:} We provide the results for the supervised linear evaluation and $K$-nn evaluation in Table \ref{table1} and Table \ref{table3} respectively, in terms of top-1 classification accuracy on the four datasets. As shown in Table \ref{table1}, with CIFAR10 dataset our method noticeably improves the accuracy of all four baselines; $1.31\%$ (BYOL), $1.34\%$ (SimCLR), $1.87\%$ (W-MSE2), and $1.42\%$ (B-Twins). With $K$-nn evaluation (without further post training) we get even more improvements, including  $2.38\%$ (BYOL), $2.4\%$ (SimCLR), $1.25\%$ (W-MSE2), and $2.15\%$ (B-Twins). Specifically the results with $K$-nn classifier clearly implies that our method upgrades the feature learning toward learning higher level features, gaining information more related to downstream task.

With CIFAR100 our method offers  $1.59\%$ and $1.92\%$ average improvements (on four baselines) respectively under supervised and $K$-nn evaluation. For Tiny ImageNet the average improvements are $1.38\%$ and $2.51\%$, respectively. %under supervised and $K$-nn evaluation. 
Out method also performs very well in upgrading the results with dataset at scale, ImageNet, as the average improvement of linear evaluation and $K$-nn evaluation on ImageNet dataset are $1.55\%$ and $2.97\%$ respectively.

\noindent\textbf{Transfer learning:} The results of transfer learning on CIFAR10/100 using ResNet50 pre-trained on ImageNet are shown in Table 3. The average improvements under supervised evaluation are $1.18\%$ and $1.31\%$ for CIFAR10 and CIFAR100, respectively. This is slightly higher than for %the average improvement in 
linear evaluation setting. 

\section{Ablation study}
% \vspace{-0.21cm}
We perform detailed ablation study on CIFAR100 
%and evaluate the framework 
under different settings.\\
\textbf{A) Baseline performance:} As presented in Tables \ref{table1}-\ref{table4}, we evaluated the performance of all four baselines without progressive supervision, with the result provided under columns "Original" for comparison purposes.\\
\textbf{B) Number of clusters:} As mentioned previously, we determine the number of clusters for $k$-means$++$ clustering based on prior knowledge for each dataset. Here, we assess the sensitivity of the framework to the number of clusters. For CIFAR100, linear evaluation with $K=100$ resulted in an average improvement (over four  baselines) of $1.59\%$, however in case of $K=50$ and $K=150$ the average improvements drop to $01.18\%$ and $0.74\%$ respectively. 
%As observed, compared to the number of classes, 
Larger $K$ tends to limit the improvement more than smaller $K$. We suspect that this could be due to the fact that smaller $K$ better encourages the compact representation of samples with the same class label, which compact representation is complementary to enlarged representation produced by sample supervision. Overall, with reasonable $K$, 
%we can see that the
progressive supervision improves the performance, as it incorporates unique component of information within the SSL training.  \\
\textbf{C) Non-progressive double supervision:} We assess the case in which the initial training continues for 800 epochs, and after generation of pseudo labels, the next phase of training continues for 200 epochs without pseudo-label update. Note that $\alpha$ is set to $10^{-1}$. Under this setting for CIFAR100, the average improvement over four baselines from $1.59\%$ (under our main setting) drops to $0.98\%$.  It seems that longer periods of pre-training under progressive supervision even with smaller $\alpha$ is more effective than short term double supervision. \\
\textbf{D) Transfer learning:} We evaluate the effectiveness of proposed approach for transfer learning scenarios, e.g. whether progressive supervision during pre-training of ImageNet would be helpful for transfer learning to CIFAR100. The result is provided in Table \ref{table4}, which demonstrates the effectiveness of framework for transfer learning scenarios. Specifically, learning the higher level features of source dataset (compared with general features of traditional SSL) is beneficial  towards classification accuracy improvement of target dataset. \\
\textbf{E) Longer pre-training:} We assess the effect of longer pre-training (on CIFAR100) only under the $K$-nn evaluation.
%in order to assess the level of learned features under longer pre-training. 
%The reason 
We use $K$-nn evaluation  to quantify improvements offered by SSL pre-training, as linear evaluation involves supervised training. With 1000, 1200, and 1500 epochs the average improvement over original performance of all baselines are respectively $1.92\%$, $2.23\%$, and $2.60\%$. 
%The performance improvement does not seem to saturate quickly with longer periods of pre-training.
\\
\textbf{F) Limitation:} 
%The limitation of this work 
One limitation is that the design of progressive supervision is mainly based on classification task as clustering for pseudo-label generation provides label for clustering. One possible direction for future research is to consider decomposition   with more than three variables. Another is on how proper data augmentation could be more aligned with SSL under PID.
%, toward the improved feature learning.  
\vspace{-0.2cm}
\section{Conclusion}
We started from the conflicting arguments on the role of mutual information in SSL, and proposed to investigate joint mutual information rather than traditional mutual information. This led to a study of SSL problem formulation within the PID framework. 
Accordingly, we propose a general pipeline which replaces the traditional single-supervisory signal with 
progressive supervision consisting of two types of supervisory signals, i.e., sample level, and cluster level supervision. The framework is versatile and readily applicable to existing baselines. Our experimental results involve detailed comparison with four different types of baselines on four publicly available datasets, and it demonstrates the effectiveness of the proposed pipeline. This work could be the beginning of a new generation of SSL baselines equipped with higher level supervision. We leave the extension of this work for segmentation and detection task for future. 

%%%%%%%%% REFERENCES
{\small
\bibliographystyle{ieee_fullname}
\bibliography{egbib}
}

\clearpage
\appendix
\section{ More on PID}
Mutual information is a fundamental concept in information theory that measures the amount of information shared between two random variables. It provides a quantitative measure of the statistical dependence or correlation between the variables. Mutual information is widely used to understand the relationship between variables and to extract relevant information.
Mathematically, the mutual information between two discrete random variables $S_1$ and $S_2$ can be defined as the average reduction in uncertainty about one variable when the other variable is known. It is represented as $I(S_1, S_2)$ and can be calculated using the following formula:

$I(S_1, S_2) = \sum\sum P(S_1, S_2) log(P(S_1, S_2) / (P(S_1) * P(S_2)))
$

Where $P(S_1)$ and $P(S_2)$ are the probability distributions of $S_1$ and $S_2$ respectively, and $P(S_1, S_2)$ is the joint probability distribution of $S_1 $and $S_2$.

However, when considering multiple variables, the concept of mutual information can be extended to joint mutual information. Joint mutual information measures the information shared between multiple variables as a whole. 

Let's consider a scenario with two source variables, $S_1$ and$ S_2$, and one target variable, $T$.
The joint mutual information between $S_1$, $S_2$, and $T$, denoted as $I(S_1, S_2; T)$, quantifies the mutual dependence of the two sources on the target. It captures how much information about the target can be obtained by knowing both $S_1$ and $S_2$ simultaneously.
PID  is a framework that further decomposes the joint mutual information into unique, redundant, and synergy components. It aims to understand the individual contributions of each source variable to the target variable.
The unique component (U) represents the information about the target that is uniquely provided by a specific source variable, independent of other sources. It captures the exclusive influence of each variable on the target.
The redundant component (R) captures the information about the target that is shared by multiple source variables (here two). It represents the overlapping contributions of the sources to the target.
The synergy component (Sy) quantifies the information that arises from the interaction between the sources when combined and is not present when considering each source individually. It captures the non-additive or nonlinear effects between the variables.
PID provides a comprehensive understanding of the relationships between variables by decomposing the joint mutual information into these distinct components. It enables the identification of unique and redundant information sources, as well as the exploration of synergistic effects among variables. Here is a typical two-source-one-target information system under PID:
\begin{equation}
    \label{eq.PID}
    % \footnotesize
         I(S_1,S_2:T)=\text{R}(T; S_1, S_2) + 
         \text{Sy}(T; S_1, S_2)+ \text{U}(T; S_1) + \text{U}(T; S_2)
\end{equation}
where $R$, $Sy$, $U$ are redundant, synergistic and unique components of information respectively.
Williams et al. \cite{williams2010nonnegative} present such decomposition in a way that all three components are non-negative. 

\begin{figure}
\label{Fig1}
  \centering
  \includegraphics[scale=.27]{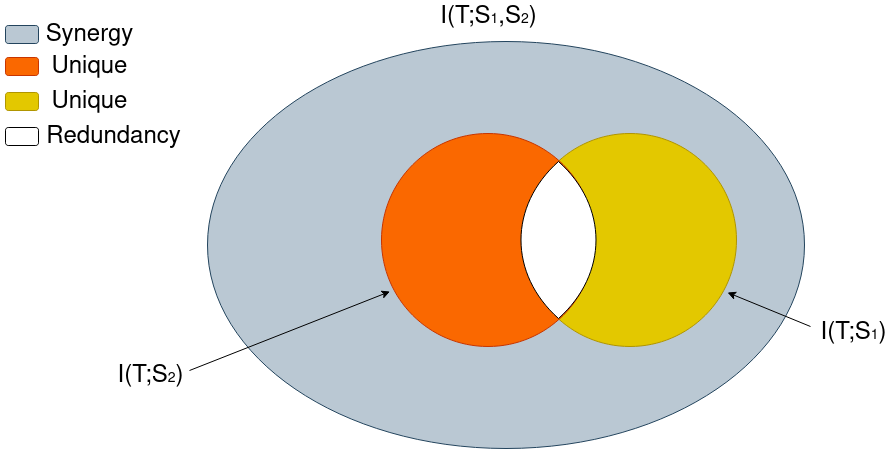}
		%\includegraphics[scale=1]{Fig1}
% 		\vspace{-0.1in}
\vspace{-1em}
		\caption{\scriptsize
		PID in case of three variables, PID presents the structure of multivariate information consisting of two source variables $S_1$ and $S_2$ as well as a target variable $T$ . 
\vspace{-3em}
}
\end{figure}

There are different ways in which one can perform the decomposition, however not every decomposition consists of non-negative components due to the confounding of redundant and synergistic interactions. \cite{williams2010nonnegative} however, presents a decomposition which all the components are non-negative. They start with a new definition for redundancy, and then leverage it to separately quantify all components as well as justifying the reason behind negative components in former decomposition methods. Fig. 1 depicts such information system with its three separately colored components.

\section{Redundant and synergistic information within SSL}
As mentioned in the former section, there are different ways to decompose joint mutual information in order to measure redundant and synergistic information in general, while not all of them lead to the separate quantification of these components. Considering a typical SSL framework that performs  feature learning via contrasting the representations of two positive views, the interaction of two representations involves both synergy and redundancy. We want to formalize such components within a SSL framework.
We consider a three variable PID system, i.e., a SSL framework that generates two views for each sample. Suppose a SSL framework generates two views ($x_1$ and $x_2$) for any given sample $x$ from a sample set, to estimate the representation of samples, $T$. The SSL framework generates representation of each view, where the corresponding representations for a sample is a random vector (or tensor in general) $V= \{V_1, V_2\}$ carrying information about the target representation $T$ as a random variable. The goal is to decompose the information provided by views' representation vector $V$ about target representation $T$, to quantify the partial information offered by subsets of $V$ ($\{V_1\}$ and $\{ V_2\}$) individually or jointly in terms of unique, redundant, and synergistic information. Roughly speaking, the SSL framework is design in such way that contrasts the representations of views and draw the representation of views from the same sample together. The information interaction between representations of views withing the SSL framework is resulted from a direct or indirect contrast of representations. Accordingly the information interaction enables the existing SSL framework to learn the target representation using only synergistic and redundant information, which are information components coming from \textbf{the two representation $V_1$ and $V_2$ and not just one of them.} The unique information from each representation however is missing within a typical SSL framework as this representation learning only learns the target representation by contrasting two other representations, not one. In fact synergistic and redundant information are components associated with two representation and not one. Now below we specify the definition of the three components of PID within SSL.

\textbf{A) Specification of Redundancy in SSL:}
If we interpret the information in SSL to be the learned features associated with each of the representations of views $x_1$ and $x_2$ ($f(x_1)$ and $f(x_2)$ with $f(.)$ being the network function), the redundancy in learned features from two representations is a measurable quantity, representing the redundant information. One way to characterize such redundant information is the correlation between the learned features from the two representations as the source variables. Note that the presumption is that we only consider the final representation in SSL to study the PID components, meaning that for the sake of simplicity we only consider the $f(.)$ and not the whole information flow within the network. This presumption is not in contradiction with the ultimate goal of SSL. In this sense, correlation between the learned features within $f(x_1)$ and $f(x_2)$ represents the redundancy. Some recent work including \cite{zbontar2021barlow,ermolov2021whitening,hua2021feature} investigate the redundancy reduction or decorrelation within SSL frameworks, and its role in better feature learning as well as avoiding the dimensional collapse. Note that they are a number of things that affect the level of correlation as well as the which features to be correlated. Data augmentation has a huge impact on that as some recent work has investigate the proper augmentation and its role \cite{tian2020makes}. However, here with SSL under PID, the presumption is that the augmentation is fixed and the analysis is only on learned features from representations using the networks. It is important to mention that with modeling SSL under PID, the goal is to decrease the redundant component of information for a better representation.

\textbf{B) Specification of Synergy in SSL:}
Taking the same approach as the one with specification of redundancy in SSL, we want to specify the synergy in feature  learning within a SSL framework. The definition of synergy in PID specifies it as the component of information that captures the non-additive or nonlinear effects between the variables, here between the representations. Accordingly we define the synergistic information component \textbf{as the components that arises from the pairwise complementariness of learned features from two representations}. For the sake of analogy, lets say a given visual feature is supposed to be learned ideally as a circle, then the representation of two views that are horizontal/vertical flip of each other allows the framework to learn the complete circle by both and not only one of them. Current SSL frameworks enjoy synergy in their respective feature learning mechanism as the augmented views of a given sample each, presents a some features in part, and to learn the complete feature, the representations of at least two positive views are required. The goal is to increase the synergy, and easiest way is to design and perform proper augmentation, in a way that views be complementary in visual features. But in terms of SSL frameworks learning mechanism, assuming the augmentation is fixed and all one can do is to design the framework, while current pipelines definitely use such component, then the goal for the framework would be to increase the exploitation of the synergistic component of information, i.e., pairwise complementariness of features, as much as possible. One way to do this is simply to define measurements of similarity in a way that it reflects the pairwise complementariness in learned representation. Accordingly, for a given Target feature $k$, one would want to have it as $f_{k}(x_1)\bigoplus f_{k}(x_2)$ where $\bigoplus$ is the direct sum operator.

\textbf{C) Specification of Unique Information in SSL:}
Easiest way to provide an example of unique information is to analyse the learning mechanism of supervised learning in classification task, where each sample has its own label and the target representation, such as a class representation is learned via samples of the class individually. However as in SSL there is no such direct supervision, the unique information extraction seems to be missing, however in other similar problem domains such as semi-supervised learning, one can say that the unique information plays an important role as the pseudo-labels make up the lack of original labels.

\section{ Synergy and redundancy in existing SSL pipelines}
While the current frameworks deal with both synergy and redundancy, as the pairwise contrasts generate such components of information in terms of learned features from the two representations, direct investigation of such components of information has been missing until recently the some work such as \cite{ermolov2021whitening,zbontar2021barlow} presented frameworks developed upon feature decorrelation for redundancy reduction. \cite{ermolov2021whitening,zbontar2021barlow} and later \cite{hua2021feature} only engaged with redundancy and not even under PID definition, however investigation of synergy component and ways to increase it, is still missing in the literature, while indirectly used in the current approaches. From another  perspective, one woudl say that while direct investigation of synergistic component of information within SSL frameworks is not available so far, however, there is a narrow literature on proper data augmentation toward reducing the common mutual information without losing the task relevant information \cite{tian2020makes},  which in essence are in favor of more exploitation of synergistic information as well as reduction of redundancy.  

Regarding the unique information, while our framework is presents the very first pipeline containing all three components of PID in a \textbf{simultaneous learning mechanism} (to the best of our knowledge), we want to confirm that some of the frameworks that utilize the SSL in conjunction of semi-supervised learning, would potentially use unique component of information even though it might not be in a simultaneous way.

\end{document}